\documentclass[sigconf]{acmart}

\acmYear{2023}
\copyrightyear{2023}
\setcopyright{rightsretained}
\acmConference[CIKM '23]{Proceedings of the 32nd ACM International Conference on Information and Knowledge Management}{October 21--25, 2023}{Birmingham, United Kingdom}
\acmBooktitle{Proceedings of the 32nd ACM International Conference on Information and Knowledge Management (CIKM '23), October 21--25, 2023, Birmingham, United Kingdom}
\acmDOI{10.1145/3583780.3614847}
\acmISBN{979-8-4007-0124-5/23/10}

\usepackage{etoolbox}
\makeatletter
\patchcmd{\maketitle}{\@copyrightpermission}{
  \begin{minipage}{0.3\columnwidth}
    \href{https://creativecommons.org/licenses/by-nc/4.0/}{\includegraphics[width=0.90\textwidth]{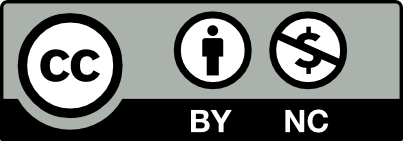}}
  \end{minipage}\hfill
  \begin{minipage}{0.7\columnwidth}
    \href{https://creativecommons.org/licenses/by-nc/4.0/}{This work is licensed under a Creative Commons Attribution-NonCommercial International 4.0 License.}
  \end{minipage}

  \vspace{5pt}
}{}{}
\usepackage{etoolbox}
\makeatletter
\patchcmd{\maketitle}{\@copyrightpermission}{
  \begin{minipage}{0.3\columnwidth}
    \href{https://creativecommons.org/licenses/by-nc/4.0/}{\includegraphics[width=0.90\textwidth]{fig/cc-by-nc4acm.png}}
  \end{minipage}\hfill
  \begin{minipage}{0.7\columnwidth}
    \href{https://creativecommons.org/licenses/by-nc/4.0/}{This work is licensed under a Creative Commons Attribution-NonCommercial International 4.0 License.}
  \end{minipage}

  \vspace{5pt}
}{}{}

\makeatother
\AtBeginDocument{%
  \providecommand\BibTeX{{%
    \normalfont B\kern-0.5em{\scshape i\kern-0.25em b}\kern-0.8em\TeX}}}

\usepackage{amsmath}
\usepackage{amssymb}
\usepackage{amsthm}
\usepackage{siunitx}
\usepackage{mathtools}
\usepackage{algorithmic}
\usepackage[lined, boxed, ruled, commentsnumbered, noend]{algorithm2e}

\usepackage{bm}
\usepackage{bbding}
\usepackage{cancel}
\usepackage{pifont}
\usepackage{enumitem}

\usepackage{float}
\usepackage{balance}
\usepackage{colortbl}
\usepackage{multirow}
\usepackage{graphicx}
\usepackage{microtype}
\usepackage{subcaption}

\usepackage[capitalize,noabbrev]{cleveref}

\definecolor{p1blue}{HTML}{3D538F}
\definecolor{p1green}{HTML}{4DA896}
\definecolor{p1yellow}{HTML}{E6B40F}
\definecolor{p1red}{HTML}{BB4430}
\definecolor{p1purple}{rgb}{0.340,0.105,0.555}

\newtheorem{problem}{Problem}

\theoremstyle{plain}

\theoremstyle{definition}

\theoremstyle{remark}

\newcommand{\R}{\mathbb{R}}

\newcommand{\identitymat}{\mathbb{I}}

\newcommand{\bigObound}{\ensuremath{\mathcal{O}}\xspace}

\newcommand{\norm}[1]{\|#1\|}

\newcommand{\round}[1]{\left(#1\right)}
\newcommand{\braces}[1]{\left\{#1\right\}}
\newcommand{\squares}[1]{\left[#1\right]}

\newcommand{\prob}[1]{\ensuremath{\text{Pr}\squares{#1}}}

\newcommand{\cikm}[1]{\textcolor{black}{{#1}}}

\newcolumntype{R}[2]{%
    >{\adjustbox{angle=#1,lap=\width-(#2)}\bgroup}%
    l%
    <{\egroup}%
}
\newcommand{\PreserveBackslash}[1]{\let\temp=\\#1\let\\=\temp}
\newcolumntype{C}[1]{>{\PreserveBackslash\centering}p{#1}}

\newcommand{\gcn}{\textsc{GCN}\xspace}
\newcommand{\sage}{\textsc{GraphSAGE}\xspace}
\newcommand{\gin}{\textsc{GIN}\xspace}

\newcommand{\pfr}{\textsc{PFR}\xspace}
\newcommand{\pfras}{\textsc{PFR-AX}\xspace}

\newcommand{\postprocess}{\textsc{PostProcess}\xspace}
\newcommand{\asim}{\textsc{EmbeddingReverser}\xspace}
\newcommand{\deepwalk}{\textsc{DeepWalk}\xspace}

\newcommand{\original}{\textsc{Original}\xspace}
\newcommand{\unaware}{\textsc{Unaware}\xspace}
\newcommand{\edits}{\textsc{EDITS}\xspace}
\newcommand{\nifty}{\textsc{NIFTY}\xspace}
\newcommand{\guide}{\textsc{GUIDE}\xspace}
\newcommand{\fairgnn}{\textsc{FairGNN}\xspace}

\newcommand{\german}{\textsc{German}\xspace}

\newcommand{\penn}{\textsc{Penn94}\xspace}

\newcommand{\pokecz}{\textsc{Pokec-z}\xspace}

\newcommand{\credit}{\textsc{Credit}\xspace}

\newcommand{\graph}{\ensuremath{\mathcal{G}}\xspace}
\newcommand{\vertexset}{\ensuremath{\mathcal{V}}\xspace}
\newcommand{\node}[1]{\ensuremath{#1}\xspace}
\newcommand{\numnodes}{\ensuremath{n}\xspace}
\newcommand{\edgeset}{\ensuremath{\mathcal{E}}\xspace}
\newcommand{\numedges}{\ensuremath{m}\xspace}
\newcommand{\A}[1]{\ensuremath{\mathbf{A}_{#1}}\xspace}
\newcommand{\adj}[2]{\ensuremath{a_{#1#2}}\xspace}
\newcommand{\diagmat}{\ensuremath{{\mathbf{D}}}\xspace}
\newcommand{\di}[1]{\ensuremath{\delta_{#1}}\xspace}
\newcommand{\neighborhood}[2]{\ensuremath{\mathcal{N}_{#2}\round{#1}}\xspace}

\newcommand{\X}{\ensuremath{\mathbf{X}}\xspace}
\newcommand{\Xhat}{\ensuremath{\mathbf{\hat{X}}}\xspace}
\newcommand{\numattributes}{\ensuremath{d}\xspace}
\newcommand{\x}[1]{\ensuremath{\mathbf{x}_{#1}}\xspace}

\newcommand{\Y}{\ensuremath{\mathbf{Y}}\xspace}
\newcommand{\y}[1]{\ensuremath{\mathbf{y}_{#1}}\xspace}
\newcommand{\Yhat}{\ensuremath{\hat{\mathbf{Y}}}\xspace}
\newcommand{\yhat}[1]{\ensuremath{\hat{\mathbf{y}}_{#1}}\xspace}
\newcommand{\Sens}{\ensuremath{\mathbf{s}}\xspace}
\newcommand{\sens}[1]{\ensuremath{s_{#1}}\xspace}

\newcommand{\U}[1]{\ensuremath{\mathbf{U}_{#1}}\xspace}
\newcommand{\Uhat}[1]{\ensuremath{\mathbf{\hat{U}}_{#1}}\xspace}

\newcommand{\embdim}{\ensuremath{k}\xspace}

\newcommand{\h}[2]{\ensuremath{\mathbf{h}_{#1}^{\round{#2}}}\xspace}
\newcommand{\numlayers}{\ensuremath{L}\xspace}
\newcommand{\model}{\ensuremath{\mathcal{M}}\xspace}
\newcommand{\logit}[1]{\ensuremath{r\round{#1}}\xspace}
\newcommand{\trainset}{\ensuremath{V}\xspace}
\newcommand{\testset}{\ensuremath{V'}\xspace}

\newcommand{\parity}{\ensuremath{\Delta_{\text{SP}}}\xspace}
\newcommand{\equality}{\ensuremath{\Delta_{\text{EO}}}\xspace}

\newcommand{\WX}[1]{\ensuremath{W^X_{#1}}\xspace}
\newcommand{\WF}[1]{\ensuremath{W^F_{#1}}\xspace}
\newcommand{\Xtilde}{\ensuremath{\mathbf{\tilde{X}}}\xspace}

\newcommand{\embN}[1]{\ensuremath{N_{\di{\node{#1}}}\round{\node{#1}}}\xspace}

\newcommand{\flipfrac}{\ensuremath{\gamma}\xspace}
\newcommand{\numtrials}{\ensuremath{T}\xspace}
\newcommand{\sensnegset}{\texttt{S1-Y0}\xspace}
\newcommand{\sensposset}{\texttt{S1-Y1}\xspace}

\begin{document}

\title{Disparity, Inequality, and Accuracy Tradeoffs in Graph Neural Networks for Node Classification}

%%%%%%%%%%%%%%%%%%%%%%%%%%%%%%%%%%%%%%%%%%%%%%%%%%%%%%%%%%%%%%%%%%%%%%%%%%%%%%%%
% CIKM
%%%%%%%%%%%%%%%%%%%%%%%%%%%%%%%%%%%%%%%%%%%%%%%%%%%%%%%%%%%%%%%%%%%%%%%%%%%%%%%%

\author{Arpit Merchant}
\orcid{0000-0001-8143-1539}
\authornote{This work was partially completed while Arpit Merchant was visiting Universitat Pompeu Fabra, Barcelona, Spain.}
\affiliation{%
  \institution{University of Helsinki}
  \streetaddress{Pietari Kalmin katu 5}
  \city{Helsinki}
  \country{Finland}
}
\email{arpit.merchant@helsinki.fi}

\author{Carlos Castillo}
\orcid{0000-0003-4544-0416}
\affiliation{%
  \institution{ICREA}
  \streetaddress{Pg.\ Llu{\'i}s Companys 23}
  \city{Barcelona}
  \country{Spain}
}
\affiliation{%
  \institution{Universitat Pompeu Fabra}
  \streetaddress{Roc Boronat 138}
  \city{Barcelona}
  \country{Spain}
}
\email{chato@icrea.cat}

\renewcommand{\shortauthors}{Merchant, et al.}

\begin{abstract}

Graph neural networks (GNNs) are increasingly used in critical human applications for predicting node labels in attributed graphs. 
Their ability to aggregate features from nodes' neighbors for accurate classification also has the capacity to exacerbate existing biases in data or to introduce new ones towards members from protected demographic groups.
Thus, it is imperative to quantify how GNNs may be biased and to what extent their harmful effects may be mitigated.
To this end, we propose two new GNN-agnostic interventions namely,
(i) \pfras which decreases the separability between nodes in protected and non-protected groups, and (ii) \postprocess which updates model predictions based on a blackbox policy to minimize differences between error rates across demographic groups.
Through a large set of experiments on four datasets, we frame the efficacies of our approaches \cikm{(and three variants)} in terms of their algorithmic fairness-accuracy tradeoff and benchmark our results against three strong baseline interventions on \cikm{three} state-of-the-art GNN models.
\cikm{Our results show that no single intervention offers a universally optimal tradeoff, but \pfras and \postprocess provide granular control and improve model confidence when correctly predicting positive outcomes for nodes in protected groups.}

\end{abstract}




\maketitle

\section{Introduction}
\label{sec:introduction}

Classification on attributed graphs involves inferring labels for nodes in the test set given a training set of labels along with attributes and adjacency information for all the nodes.
To address this task, Graph Neural Networks (or GNNs, for short) have exploded in popularity since they effectively combine attributes and adjacency to build a unified node representation which can be used downstream as a feature vector~\cite{Hamilton2017,xiao2022graphSurvey}.
GNNs have found applications in a variety of high-risk application domains (as defined, e.g., in the proposed AI Act for Europe of April 2022\footnote{URL: \url{https://artificialintelligenceact.eu/} (retrieved January 2023).}), including credit risk applications~\cite{dua2019uci}, and crime forecasting~\cite{jin2020crime}.
Here, nodes usually represent individuals, and node attributes include sensitive information indicating membership in demographic groups protected by anti-discrimination regulations.
In such cases, we ideally want algorithmic models to predict labels accurately while ensuring that the predicted labels do not introduce a systematic disadvantage for people from protected groups.
For instance, in the case of risk assessment for credit, models should correctly infer whether a client is likely to repay a loan, and should not introduce an unwanted bias against applicants because of their national origin, age, gender, or other protected attribute.
An entire field has been devoted in recent years to these \emph{algorithmic discrimination} concerns~\cite{kleinberg2019discrimination,raghavan2020mitigate}.

A key challenge in making predictions that are algorithmically fair arises from the multimodal nature of graph data, i.e., attributes and adjacency.
Unlike traditional machine learning~\cite{zemel2013learning}, delinking the correlations of sensitive attributes to other attributes is insufficient; proximity to other nodes in the same protected group can indirectly indicate membership and this may propagate into node representations.
Thus, reducing bias may additionally require learning to deemphasize correlations in adjacency information.
While numerous GNN architectures have been proposed to achieve state-of-the-art accuracy on different datasets~\cite{Li2018,graphsaint-iclr20}, recent studies show that they may algorithmically discriminate due to their tendency to exacerbate existing biases in data or introduce new ones during training~\cite{dong2022edits}. 
This has motivated the design of GNN-agnostic methods such as \edits~\cite{dong2022edits}, which adversarially modifies graph data via an objective function that penalizes bias and \nifty~\cite{agarwal2021towards}, which augments a GNN's training objective through layer-wise weight normalization to jointly reduce bias and improve stability.

However, such interventions from previous literature differ in numerous ways making meaningful comparisons of their respective efficacies difficult.
First, different methods adopt different frameworks and may optimize different metrics (e.g., \edits uses Wasserstein and reachability distances~\cite{dong2022edits}, \nifty uses counterfactual unfairness~\cite{agarwal2021towards}, \cikm{\guide uses a group-equality informed individual fairness criteria~\cite{song2022guide}}).
Second, dataset properties, training criteria, hyperparameter tuning procedures, and sometimes, even low-level elements of an implementation such as linked libraries are known to significantly influence the efficiency and effectiveness of GNNs on node classification~\cite{zhu2020beyond,gao2020gnnsearch}. 
Third, while algorithmic discrimination may be reduced at the expense of accuracy~\cite{liang2022design}, specific improvements and trade-offs depend on application contexts~\cite{pmlr-v81-menon18a}, and need to be evaluated to understand what kinds of alternatives may offer improvements over current approaches.
\cikm{Our goal is to address these limitations by focusing on the following questions:
\begin{itemize}[leftmargin=10mm]
	\item[\textbf{RQ1}:] How do we meaningfully benchmark and analyze the tradeoff between algorithmic fairness and accuracy of interventions on GNNs across different graphs?
	\item[\textbf{RQ2}:] Is there room for improving the fairness/accuracy tradeoff, and if so, how?
\end{itemize}
}

\paragraph{Our Contributions.}
We categorize interventions designed to reduce algorithmic discrimination in terms of their loci in the machine learning pipeline: 
 (a) pre-processing, before training, (b) in-processing, during learning, and (c) post-processing, during inference.
\cikm{Using a standardized methodological setup, we seek to maximally preserve accuracy while improving algorithmic fairness.
To this end,} we introduce two new, unsupervised (independent of ground-truth labels), model-agnostic \cikm{(independent of the underlying GNN architecture)} interventions; \pfras that debiases data prior to training, and \postprocess that debiases model outputs after training (before issuing final predictions). 
% \cikm{Both these interventions can be applied to any GNN model.}

In \pfras, we first use the \pfr method~\cite{lahoti2019pfr} to transform node attributes to better capture data-driven similarity for operationalizing individual fairness.
Then, we construct a DeepWalk embedding~\cite{perozzi2014deepwalk} of the graph, compute its \pfr transformation, and reconstruct a graph from the debiased embedding using a method we call \asim. 
\cikm{To our knowledge, this is a novel application of a previously known method with suitable augmentations.}

In \postprocess, we randomly select a small fraction, referred to as \flipfrac, of nodes from the minority demographic for whom the model has predicted a negative outcome and update the prediction to a positive outcome.
This black-box policy aims to ensure that error rates of a model are similar across demographic groups.
This is a simple and natural post-processing strategy which, to the best of our knowledge, has not been studied in the literature on GNNs.

We conduct extensive experiments to evaluate the efficacies of interventions grouped by their aforementioned loci.
To measure accuracy, we use \emph{AUC-ROC}; to measure algorithmic fairness, we use \emph{disparity} and \emph{inequality} (cf. \cref{sec:problem_setup}).
We compare the accuracy-fairness tradeoff for \pfras and \postprocess \cikm{(plus three additional variants)} against three powerful baseline interventions (two for pre-training, one for in-training) on \cikm{three} widely used GNN models namely, \gcn, \sage, and \gin~\cite{xu2018powerful}.
\cikm{Our experiments are performed on two semi-synthetic and two real-world datasets with varying levels of edge homophily with respect to labels and sensitive attributes}, which is a key driver of accuracy and algorithmic fairness in the studied scenarios.
We design ablation studies to measure the effect of the components of \pfras and the sensitivity of \postprocess to the \flipfrac parameter.
\cikm{Finally, we analyze the impact of interventions on model confidence.}
Our main findings are summarized below:
\begin{itemize}
  \item \cikm{No single intervention offers universally optimal tradeoff across models and datasets.}
  \item \pfras and \postprocess provide granular control over the accuracy-fairness tradeoff compared to baselines. Further, they serve to improve model confidence in correctly predicting positive outcomes for nodes in protected groups.
  \item \pfr-A and \pfr-X that debias only adjacency and only attributes respectively, offer steeper tradeoffs than \pfras which debiases both. 
  \item When imbalance between protected and non-protected groups and model bias are both large, small values of \flipfrac offer large benefits to \postprocess.
\end{itemize}

\section{Related Work}
\label{sec:related_work}

Legal doctrines such as GDPR (in Europe), the Civil Rights Act (in the US), or IPC Section 153A (in India) restrict decision-making on the basis of protected characteristics such as nationality, gender, caste~\cite{stephanopoulos2018disparate}.
While \emph{direct discrimination}, i.e., when an outcome directly depends on a protected characteristic, may be qualitatively reversed, addressing \emph{indirect discrimination}, i.e., discrimination brought by apparently neutral provisions, requires that we define concrete, quantifiable metrics in the case of machine learning (ML) that can be then be optimized for~\cite{zemel2013learning}.
Numerous notions of algorithmic fairness have been proposed and studied~\cite{hardt2016equality}.
Two widely used definitions include the \emph{separation criteria}, which requires that some of the ratios of correct/incorrect positive/negative outcomes across groups are equal, and the \emph{independence criterion}, which state that outcomes should be completely independent from the protected characteristic~\cite{barocas2016big}. 

\paragraph{Algorithmic Fairness-Accuracy Tradeoffs.}
However, including fairness constraints often results in classifiers having lower accuracy than those aimed solely at maximizing accuracy. 
Traditional ML literature~\cite{kleinberg2019discrimination,zafar2017mistreatment} has extensively studied the inherent tension that exists between technical definitions of fairness and accuracy:
\citet{davies2017cost} theoretically analyze the cost of enforcing disparate impact on the efficacy of decision rules;
\citet{lipton2018mitigate} explore how correlations between sensitive and nonsensitive features induces within-class discrimination;
\citet{fish2016confidence} study the resilience of model performance to random bias in data.
In turn, characterizing these tradeoffs has influenced the design of mitigation strategies and benchmarking of their utility.
Algorithmic interventions such as reweighting training samples~\cite{kamiran}, regularizing training objectives to dissociate outcomes from protected attributes~\cite{liu2019incorporating}, and adversarially perturbing learned representations to remove sensitive information~\cite{elazar2018adversarial} are framed by their ability to reduce bias without significantly compromising accuracy.

\paragraph{Algorithmic Fairness in GNNs.}
The aforementioned approaches are not directly applicable for graph data due to the availability of adjacency information and the structural and linking bias it may contain. 
GNNs, given their message-passing architectures, are particularly susceptible to exacerbating this bias.
This has prompted attention towards mitigation strategies for GNNs.
\cikm{For instance, at the pre-training phase, REDRESS~\cite{dong2021redress} seeks to promote individual fairness for the ranking task, and at the in-training phase, \fairgnn~\cite{dai2021fairgnn} estimates missing protected attribute values for nodes using a GCN-estimator for adversarial debiasing and \guide~\cite{song2022guide} proposes a novel GNN model for a new group-equality preserving individual fairness metric.
We do not compare against these since they are designed for a different task than ours, operate in different settings altogether, and since \fairgnn (in particular) exhibits a limited circular dependency on using vanilla GNN for a sensitive task to overcome limitations of a different GNN for classification.}
We refer the reader to \citet{dai2022comprehensive} for a recent survey.
More relevant to our task, EDITS~\cite{dong2022edits} reduces attribute and structural bias using a Wasserstein metric and so we use it as a baseline for comparison.
At the in-training phase, NIFTY~\cite{agarwal2021towards} promotes a model-agnostic fair training framework for any GNN using Lipschitz enhanced message-passing.
However, an explicit fairness-accuracy tradeoff analysis is lacking from literature which, along with methodological differences, makes it difficult to benchmark the comparative utilities of these approaches. 
Therefore, we include these as baselines.
We frame our study in the context of such an analysis and design one pre-training and one post-training intervention that offer different, but useful tradeoffs.

\section{Problem Setup}
\label{sec:problem_setup}

\paragraph{Graphs.}
Let $\graph = \round{\vertexset, \edgeset}$ be an unweighted, undirected graph where \vertexset is a set of \numnodes nodes and \edgeset is a set of \numedges edges. 
Denote $\A{} = \squares{\adj{\node{u}}{\node{v}}} \in \braces{0,1}^{\numnodes \times \numnodes}$ as its binary adjacency matrix where each element \adj{\node{u}}{\node{v}} indicates the presence or absence of an edge between nodes \node{u} and \node{v}.
Define $\diagmat = \text{diag} \round{\di{1}, \di{2},\ldots,\di{\numnodes}}$ to be a diagonal degree matrix where $\di{\node{u}} = \sum_{\node{v}} \adj{\node{u}}{\node{v}}$.
Let each node \node{u} in \graph be associated with one binary sensitive attribute variable \sens{\node{u}} indicating membership in a protected demographic group along with $\numattributes-1$ additional real or integer-valued attributes.
Together, in matrix form, we denote node attributes as $\X \in \R^{\numnodes \times \numattributes}$.
Lastly, $\forall \node{u} \in \vertexset$, its binary, ground-truth, categorical label is depicted as $\y{\node{u}}$.

\paragraph{Graph Neural Networks.}
Typically, GNNs comprise of multiple, stacked graph filtering and non-linear activation layers that leverage \X and \A{} to learn joint node representations (see, e.g.,~\citet{Kipf2016}).
Such a GNN with \numlayers layers captures the \numlayers-hop neighborhood information around nodes. 
For each $\node{v} \in \vertexset$ and $l \in \squares{\numlayers}$, let \h{\node{v}}{l} denote the representation of node \node{v} at the $l$-th GNN layer.
In general, \h{\node{v}}{l} is formulated via message-passing as follows:
\begin{equation}\label{eq:general_gnn_node_repr}
	\small{\h{\node{v}}{l} = \text{CB}^{\round{l}} \round{\h{\node{v}}{l-1}, \text{AGG}^{\round{l-1}}\round{\braces{\h{v}{l-1}: \node{u} \in \neighborhood{\node{v}}{}}}}}
\end{equation}
where \neighborhood{\node{v}}{} is the neighborhood of \node{v}, \h{\node{v}}{l-1} is the representation of \node{v} at the $\round{l-1}$-th layer, AGG is an aggregation operator that accepts an arbitrary number of inputs, i.e., messages from neighbors, and CB is a function governing how nodes update their representations at the $l$-th layer. 
At the input layer, \h{\node{v}}{0} is simply the node attribute $\x{\node{v}} \in \X$ and \h{\node{v}}{\numlayers} is the final representation. 
Finally, applying the softmax activation function on \h{\node{v}}{\numlayers} and evaluating cross-entropy error over labeled examples, we can obtain predictions for unknown labels \yhat{\node{v}}.
In this paper, we use AUC-ROC and F1-scores (thresholded at 0) to measure GNN accuracy.

\paragraph{Algorithmic Fairness.}
We measure the algorithmic fairness of a GNN model using two metrics.
First, \emph{Statistical Disparity} (\parity), based on the \emph{independence criterion}, captures the difference between the positive prediction rates between members in the protected and non-protected groups~\cite{dwork2012parity}. 
Formally, for a set of predicted labels \Yhat:
\begin{equation}\label{eq:parity_definition}
	\small{\parity = \biggm\vert \prob{\Yhat = 1 | \Sens = 1} - \prob{\Yhat = 1 | \Sens = 0}\biggm\vert}
\end{equation}
Second, \emph{Inequal Opportunity} (\equality), which is one \emph{separation criterion}, measures the similarity of the true positive rate of a model across groups~\cite{hardt2016equality}. 
Formally:
\begin{equation}\label{eq:equality_definition}
	\small{\equality = \biggm\vert \prob{\Yhat = 1 | \Sens = 1, \Y = 1} - \prob{\Yhat = 1 | \Sens = 0, \Y = 1}\biggm\vert}
\end{equation}
\cref{eq:equality_definition} compares the probability of a sample with a positive ground-truth label being assigned a positive prediction across sensitive and non-sensitive groups. 
%
% \inote{In the following we call \parity disparity and \equality inequality to emphasize that lower values are better.}
%
In the following sections, we refer to \parity as \emph{disparity} and \equality as \emph{inequality} to emphasize that lower values are better since they indicate similar rates. 

Having defined the various elements in our setting, we formally state our task below:
\begin{problem}[Algorithmically Fair Node Classification]\label{prob:node_classification}
    Given a graph \graph as an adjacency matrix $\A{}$, node features $\X$ including sensitive attributes \Sens, and labels $\Y_{V}$ for a subset of nodes $V \subset \vertexset$, debias GNNs such that their predicted labels $\Y_{\vertexset \setminus V}$ are maximally accurate while having low \parity and \equality.
\end{problem}

\section{Algorithms}
\label{sec:algorithms}

In this section, we propose two algorithms for Problem~\ref{prob:node_classification}: \pfras (pre-training) and \postprocess (post-training).

\subsection{\pfras}\label{subsec:pfras}

Our motivation for a data debiasing intervention arises from recent results showing that GNNs have a tendency to exacerbate homophily~\cite{zhu2020beyond}.
Final node representations obtained from GNNs homogenize attributes via Laplacian smoothing based on adjacency. 
This has contributed to their success in terms of classification accuracy.
However, it has also led to inconsistent results for nodes in the protected class when their membership status is enhanced in their representations due to message-passing~\cite{dong2022edits,khajehnejad2022crosswalk}, particularly in cases of high homophily.
\citet{lahoti2019pfr} design \pfr to transform attributes to learn new representations that retain as much of the original data as possible while mapping equally deserving individuals as closely as possible.
The key benefit offered by \pfr is that it obfuscates protected group membership by reducing their separability from points in the non-protected group.
Therefore, we directly adapt \pfr for graph data to debias attributes and adjacency.
Algorithm~\ref{algo:pfras} presents the pseudocode for \pfras.

\begin{algorithm}[tb]

	\begin{algorithmic}

		\STATE {\bfseries Input:} Graph $\graph = \round{\vertexset, \edgeset}$ as adjacency matrix \A{}; Degree matrix \diagmat; Node attributes \X; Sensitive attributes \Sens; Ranking variable $Z$; Number of rounds $T_{\text{SC}}$;
		\STATE {\bfseries Output:} Debiased attributes \Xhat; Debiased graph $\tilde{\graph }$;

		\STATE \textcolor{purple}{\slash* Debias Attributes *\slash}
		\STATE $\Xhat \gets$ \pfr$\round{\X, Z, \Sens}$ \hspace*{0pt}\hfill $\ldots$Solve Equation~\ref{eq:pfr_optimization}

		\STATE \textcolor{purple}{\slash* Debias Adjacency *\slash}
		\STATE $\U{} \gets$ \deepwalk$\round{\A{}}$ \hspace*{0pt}\hfill $\ldots$cf. Equation~\ref{eq:deepwalk_factorization}
		\STATE $\Uhat{} \gets$ \pfr$\round{\U, Z, \Sens}$ \hspace*{0pt}\hfill $\ldots$Solve Equation~\ref{eq:pfr_optimization}

		\STATE \textcolor{purple}{$\vartriangleright$ \asim}
		\STATE $M \gets 0$
		\STATE $\tilde{\graph} \gets \text{Initialize empty graph}$
		\FOR{$\node{u}$ {\bfseries in} \vertexset}
		\STATE $d\squares{\node{u}} \gets 0$, ~ $\text{completed}\squares{\node{u}} \gets False$
		\ENDFOR

		\FOR{$t$ {\bfseries in} $T_{\text{SC}}$ rounds}
			\FOR{\node{u} {\bfseries in} \vertexset}
				\IF{$\text{completed}\squares{\node{u}}$ is $False$}
					\STATE $\embN{u} \gets \di{\node{u}}$ nearest neighbors of \node{u} in \Uhat{}
					
					\FOR{\node{v} {\bfseries in} \embN{u}}
						\STATE $\embN{v} \gets \di{\node{v}}$ nearest neighbors of \node{v} in \Uhat{}

						\IF{$\text{completed}\squares{\node{j}}$ is $False$ and $\node{u} \in \embN{v}$}
							\STATE Add edge $\round{\node{u},\node{v}}$ to $\tilde{\graph}$
							\STATE Increment $M$
							\STATE \textbf{if} $M \geq |\edgeset|$, \textbf{break}
							\STATE \textbf{if} $d\squares{\node{u}} \geq \di{\node{u}}$, \textbf{then} $\text{completed}\squares{\node{u}} \gets True$
							\STATE \textbf{if} $d\squares{\node{v}} \geq \di{\node{v}}$, \textbf{then} $\text{completed}\squares{\node{v}} \gets True$

						\ENDIF
					\ENDFOR
				\ENDIF
			\ENDFOR
		\ENDFOR

	\end{algorithmic}
	 	
	\caption{\pfras}
	\label{algo:pfras}
	
\end{algorithm}

\paragraph{Debiasing Attributes.}
In order to transform attributes \X using \pfr, we build two matrices.
The first, denoted by \WX{}, is an adjacency matrix corresponding to a $k$-nearest neighbor graph over \X (not including \Sens) and is given as:
\begin{equation}\label{eq:pfr_knn_graph_matrix}
	\WX{\node{u}\node{v}} = 
	\begin{cases}
		\small{\exp\round{\frac{-\norm{\x{\node{u}}-\x{\node{v}}}^2}{t}}, ~\text{if}~ \node{u} \in N_k\round{\node{v}} ~\text{or}~ \node{v} \in N_k\round{\node{u}}} \\
		0, ~\text{otherwise}
	\end{cases}
\end{equation}
where $t$ is a scaling hyperparameter and $N_k\round{\node{v}}$ is the set of $k$ nearest neighbors of $\node{v}$ in Euclidean space. 
We first normalize \X using Min-Max scaling to ensure that all attributes contribute equally and then compute \WX{} as per Equation~\ref{eq:pfr_knn_graph_matrix}.
The second matrix, denoted by \WF{}, is the adjacency matrix of a between-group quantile graph that ranks nodes within protected and non-protected groups separately based on certain pre-selected variables and connects similarly ranked nodes.
In the original paper, \citet{lahoti2019pfr} use proprietary decile scores obtained from Northpointe for creating rankings.
However, in the absence of such scores for our data, we use one directly relevant attribute for the task at hand.
For instance, in the case of a credit risk application, we define rankings based on the loan amount requested.
Formally, this matrix is given as:
\begin{equation}\label{eq:pfr_between_quantile_graph_matrix}
	\WF{\node{u}\node{v}} = 
	\begin{cases}
		\small{1, ~\text{if}~ \node{u} \in X_{\sens{\node{u}}}^p ~\text{and}~ \node{v} \in X_{\sens{\node{v}}}^p }, ~ \sens{\node{u}} \neq \sens{\node{v}} \\
		0, ~\text{otherwise}
	\end{cases}
\end{equation}
where $X_s^p$ denotes the subset of nodes with sensitive attribute value $s$ whose scores lie in the $p$-th quantile. 
Higher number of quantiles leads a sparser \WF{}.
Thus, \WF{} is a multipartite fairness graph that seeks to build connections between nodes with different sensitive attributes based on similarity of their characteristics even if they are not adjacent in the original graph.
Finally, a new representation of \X, denoted as \Xtilde, is computed by solving the following problem~\cite{lahoti2019pfr}:
\begin{equation}\label{eq:pfr_optimization}
	\begin{aligned}
		\text{minimize}_{\tilde{X}}       \quad & \round{1-\alpha} \sum\limits_{\node{u},\node{v}}^{\numnodes} \norm{\tilde{x}_{\node{u}} - \tilde{x}_{\node{v}}}^2 \WX{\node{u}\node{v}} \\
		& \quad \quad \quad \quad + \alpha \sum\limits_{\node{u},\node{v}}^{\numnodes} \norm{\tilde{x}_{\node{u}} - \tilde{x}_{\node{v}}}^2 \WF{\node{u}\node{v}} \\
		\textnormal{s.t.} \quad & \tilde{X}^{\top}\tilde{X} = \identitymat \\
	\end{aligned}
\end{equation}
where $\alpha$ controls the influence of \WX{} and \WF{} on \Xtilde.

\paragraph{Debiasing Adjacency.}
To reduce linking bias from \A{}, we apply a three-step process. 
First, we compute an unsupervised node embedding of the graph using a popular matrix factorization approach named DeepWalk~\cite{perozzi2014deepwalk}. Formally, this is computed as follows:
\begin{equation}\label{eq:deepwalk_factorization}
	\U{} = \log\round{\text{vol}\round{\graph}\round{\frac{1}{C}\sum_{c=1}^C \round{\diagmat^{-1}\A{}}^c} \diagmat^{-1}} - \log b
\end{equation}
where $\text{vol}\round{\graph} = 2\numedges/\numnodes$ is the volume of the graph, $C$ represents the length of the random walk, and $b$ is a hyperparameter controlling the number of negative samples.
Second, using the same aforementioned procedure for debiasing \X, we apply \pfr on \U{}.
Third, we design a new algorithm to invert this debiased embedding to reconstruct a graph with increased connectivity between nodes in protected and non-protected groups. 
This algorithm, which we refer to as \asim, proceeds as follows.
We initialize an empty graph of \numnodes nodes and locate for each node \node{u}, its \di{\node{u}} closest neighbors in the embedding space denoted as \embN{u} where \di{\node{u}} is the degree of \node{u} in the original graph. 
Starting from the first node (say) \node{v}, for every $\node{w} \in \embN{v}$, we check if \node{v} is present in \node{w}'s \di{\node{w}} closest neighbors.
If so, we add an edge between \node{v} and \node{w} and increment counters corresponding to the current degrees for \node{v} and \node{w}.
We also increment a global counter maintaining the number edges added so far.
If the current degree for any node (say) \node{u} reaches \di{\node{u}}, we mark that node as completed and remove it from future consideration.
This continues either for $T_{\text{SC}}$ rounds where each round iterates over all nodes or until \numedges edges have been added. 
Thus, we seek to maximally preserve the original degree distribution.

\begin{algorithm}[tb]

	\begin{algorithmic}

		\STATE {\bfseries Input:} Test set \testset; Sensitive attribute values $\Sens{}_{\testset}$; Model predictions $\Yhat_{\testset}$; Model output scores $\logit{\cdot}$ for \testset; Flip parameter \flipfrac; confidence (uncalibrated) \texttt{MAX-SCORE};
		\STATE {\bfseries Output:} Updated model predictions $\Yhat_{\testset}$; Updated model output scores $\logit{\cdot}$;

		\STATE \texttt{S1-Y0} $\gets \emptyset$ % \COMMENT{Identify nodes in protected group with negative predicted outcome}

		\FOR{$\node{u}$ {\bfseries in} \testset}
		\IF{$\sens{\node{u}} = 1$ and $\yhat{\node{u}} = 0$}
		\STATE \texttt{S1-Y0} $\gets$ \texttt{S1-Y0} $\cup ~\braces{\node{u}}$
		\ENDIF
		\ENDFOR

		\STATE $P \gets$ Randomly select \flipfrac fraction of nodes from \texttt{S1-Y0}
		\FOR{$\node{v}$ {\bfseries in} $P$}
		\STATE $\yhat{\node{v}} \gets 1$
		\STATE $\logit{\node{v}} \gets \texttt{MAX-SCORE}$
		\ENDFOR

	\end{algorithmic}
	 	
	\caption{\postprocess}
	\label{algo:postprocess}
	
\end{algorithm}

\subsection{\postprocess}\label{subsec:postprocess}

\paragraph{Model Predictions.}
Let \model be a GNN model trained on a set of nodes $\trainset \in \vertexset$.
Let $\testset = \vertexset \setminus \trainset $ represent nodes in the test set and let $\Sens_{\testset}$ be their sensitive attribute values. 
For any $\node{u} \in \testset$, denote $\logit{\node{u}} \in \R$ as the original output (logit) score capturing the uncalibrated confidence of \model.
In our binary classification setting, we threshold $\logit{\cdot}$ at 0 and predict a positive outcome for \node{u}, i.e. $\yhat{\node{u}} = 1$, if $\logit{\node{u}} \geq 0$.
Otherwise, we predict a negative outcome.
Denote $\Yhat_{\testset}$ as the set of labels predicted by \model. 

\paragraph{Do-No-Harm Policy.}
Next, we present our model-agnostic post-training intervention called \postprocess which operates in an unsupervised fashion independent of ground-truth labels.
Different from prior interventions, especially~\citet{pmlr-v108-wei20a}, \postprocess seeks to relabel model predictions following a \emph{do-no-harm policy}, in which protected nodes with a positive outcome are never relabeled to a negative outcome.
We audit $\Yhat_{\testset}$ and $\Sens_{\testset}$ to identify all the nodes in the test set belonging to the protected class ($\sens{} = 1$) that have been assigned a negative outcome ($\yhat{} = 0$).
Denote this set as \sensnegset (and so on for \sensposset, etc.).
For a fixed parameter $\flipfrac \in \squares{0,1}$, we randomly select a \flipfrac fraction of nodes from \sensnegset and change their predicted label to a positive outcome, i.e., $\yhat{} = 1$.
Then, we update \model's scores for these nodes to a sufficiently large (uncalibrated) positive value. 
That is, we post-process \model to be confident about its new predicted labels.
Predictions for all other nodes in the test set remain unchanged.
Algorithm~\ref{algo:postprocess} describes the pseudocode.

\paragraph{Choice of \flipfrac.}
Determining a useful value for \flipfrac depends on two factors: (i) imbalance in the test set with respect to the number of nodes in the protected class, and (ii) bias in \model's predictions towards predicting negative outcomes.
If imbalance and bias are large, small \flipfrac values may be sufficient to reduce disparity. 
If imbalance is low and bias is large, then large \flipfrac values may be required.
\cikm{Let $\hat{n}_{\sensposset}$ denote the number of nodes in \sensposset, and similarly for \sensnegset, etc. Then, disparity (Equation~\ref{eq:parity_definition}) is rewritten as:}
\begin{equation*}
	\parity = \biggm\vert \frac{\hat{n}_{\sensposset}}{\hat{n}_{\sensposset} + \hat{n}_{\sensnegset}} - \frac{\hat{n}_{\texttt{S0-Y1}}}{\hat{n}_{\texttt{S0-Y1}} + \hat{n}_{\texttt{S0-Y0}}} \biggm\vert
\end{equation*}

\cikm{Our do-no-harm policy reduces $\hat{n}_{\sensnegset}$ and increases $\hat{n}_{\sensposset}$. $\hat{n}_{\sensposset} + \hat{n}_{\sensnegset}$ remains constant. Thus, the first term in the equation above increases while the second remains the same.
If the difference between the first and second terms is small, then \postprocess will increase disparity. Conversely, if the difference is large, then \postprocess will reduce disparity. If $\hat{n}_{\sensposset} >> \hat{n}_{\sensnegset}$, then \postprocess will have marginal impact on disparity. The effect on \equality follows equivalently, but may not be correlated with \parity.}

Note, the impact of \flipfrac on accuracy cannot be determined due to the unavailability of ground-truth label information during this phase.
So, in Section~\ref{subsec:experiment_results}, we empirically analyze the impact of \flipfrac on accuracy, averaged over \numtrials trials for smoothening.

\section{Experiments}
\label{sec:experiments}

\cikm{In this section, we describe the datasets and the methodology used in our experimental study and report our findings.}
% We empirically compare our algorithms with four baselines on three GNN models over four datasets.

\begin{table}[t!]
	\caption[caption]{Dataset Statistics: number of nodes ($|\mathcal{V}|$), number of edges ($|\mathcal{E}|$), sensitive attribute \textbf{$s$}, \cikm{label \textbf{$l$}}, sensitive attribute homophily\footnotemark~ ($h_s$), label homophily ($h_l$).}

	\fontsize{9}{12}\selectfont 

	\centering 

	\begin{tabular}{lcccccc} 

		\toprule 

		\multirow{2}[2]{*}{\textbf{Dataset}} & \multicolumn{2}{c}{\textbf{Size}} & \multicolumn{4}{c}{\textbf{Properties}} \\ 

		\cmidrule(lr){2-3} \cmidrule(lr){4-7} 

		& $|\mathcal{V}|$ & $|\mathcal{E}|$ & \textbf{$s$} & \cikm{\textbf{$l$}} & $h_s$ & $h_l$ \\

		\midrule

		\textsc{German} &  1K &   21K & Gender & \cikm{Good Risk} & 0.81 & 0.60 \\
		\textsc{Credit} & 30K & 1.42M &    Age & \cikm{No Default} & 0.96 & 0.74 \\
		\textsc{Penn94} & 41K & 1.36M & Gender & \cikm{Year} & 0.52 & 0.78 \\
		\textsc{Pokec-z} & 67K &  617K & Region & \cikm{Profession} & 0.95 & 0.74 \\

		\bottomrule 

	\end{tabular} 

	\label{tab:dataset-statistics}

\end{table}

\subsection{Datasets}\label{subsec:datasets}

\cikm{We evaluate our interventions on four publicly-available datasets ranging in size from 1K to 67K nodes.
For consistency, we binarize sensitive attributes ($\sens{}$) and labels in each dataset.
$\sens{} = 1$ indicates membership in the protected class and 0 indicates membership in the non-protected class. 
Similarly, label values set to 1 indicate a positive outcome and 0 indicate a negative outcome.
Table~\ref{tab:dataset-statistics} presents a summary of dataset statistics.\footnotetext{Homophily in relation to an attribute is defined as the ratio of number of edges with both end-points having the same value to the total number of edges.}}

\paragraph{Semi-Synthetic Data.}
\german~\cite{dua2019uci} consists of clients of a German bank where the task is to predict whether a client has good or bad risk independent of their \emph{gender}. 
\credit~\cite{ustun2019credit} comprises of credit card users and the task is to predict whether a user will default on their payments.
Here, \emph{age} is the sensitive attribute.
Edges are constructed based on similarity between credit accounts (for \german) and purchasing patterns (for \credit), following \citet{agarwal2021towards}.
We add an edge between two nodes if the similarity coefficient between their attribute vectors is larger than a pre-specified threshold. 
This threshold is set to 0.8 for \german and 0.7 for \credit.

\paragraph{Real-world Data.}
\cikm{In \penn~\cite{traud2012penn}, nodes are Facebook users, edges indicate friendship, and the task is to predict the graduation year~\cite{lim2021large} independent of \emph{gender} (sensitive attribute).
\pokecz~\cite{dai2021fairgnn} is a social network of users from Slovakia where edges denote friendship, \emph{region} is a sensitive attribute, and labels indicate professions.}

\begin{figure*}[ht]
	\captionsetup{skip=3pt}
	\captionsetup[sub]{skip=3pt}
	\centering
	\includegraphics[width=0.95\textwidth]{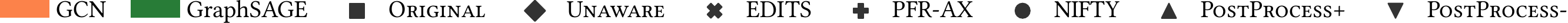}
	\\
	\begin{subfigure}{0.23\textwidth}
	  \centering
	  \includegraphics[width=\textwidth]{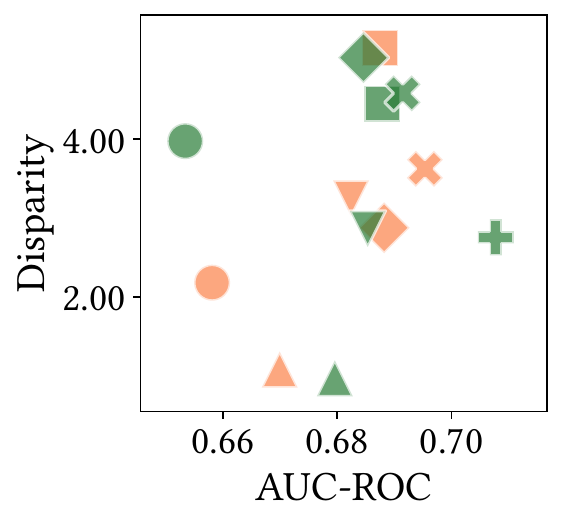}
	  \label{fig:German_AUC-ROC_Parity}
	\end{subfigure}
	\hspace{1mm}
	\begin{subfigure}{0.23\textwidth}
		\centering
		\includegraphics[width=\textwidth]{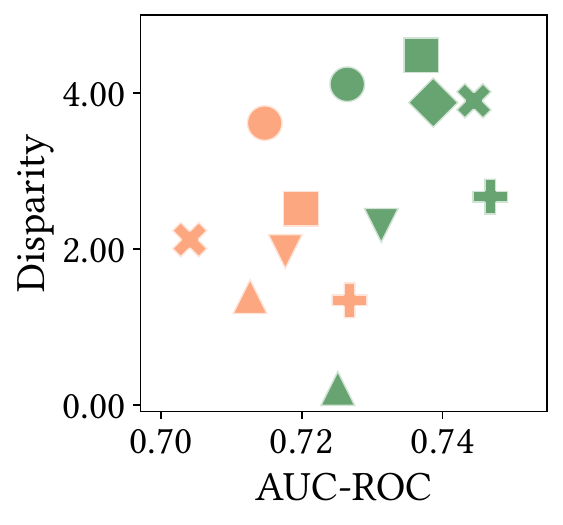}
		\label{fig:Credit_AUC-ROC_Parity}
	\end{subfigure}
	\hspace{1mm}
	\begin{subfigure}{0.23\textwidth}
		\centering
		\includegraphics[width=\textwidth]{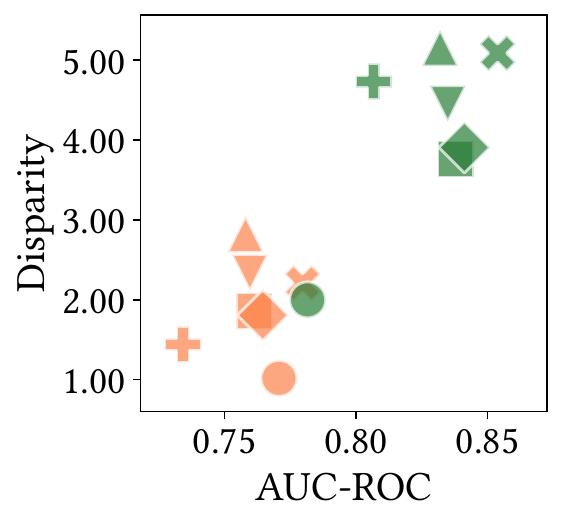}
		\label{fig:Penn94_AUC-ROC_Parity}
	\end{subfigure}
	\hspace{1mm}
	\begin{subfigure}{0.23\textwidth}
		\centering
		\includegraphics[width=\textwidth]{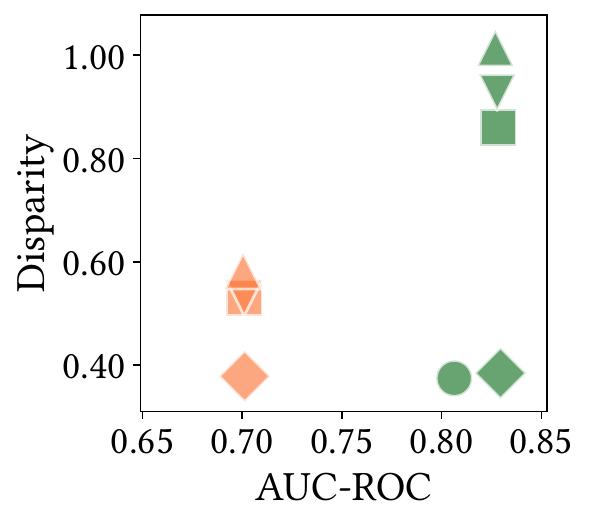}
		\label{fig:Pokec-Z_AUC-ROC_Parity}
	\end{subfigure}
	\begin{subfigure}{0.23\textwidth}
		\centering
		\includegraphics[width=\textwidth]{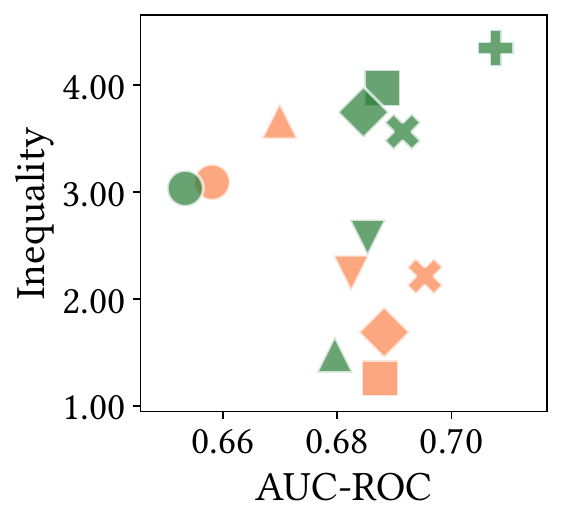}
		\caption{\german}
		\label{fig:German_AUC-ROC_Equality}
	  \end{subfigure}
	  \hspace{1mm}
	  \begin{subfigure}{0.23\textwidth}
		\centering
		\includegraphics[width=\textwidth]{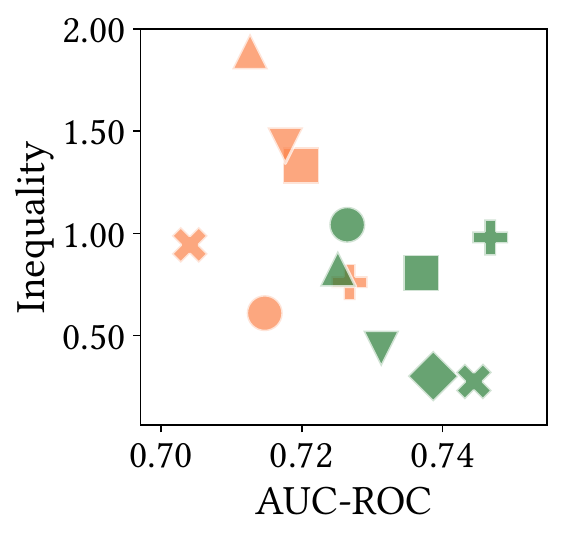}
		\caption{\credit}
		\label{fig:Credit_AUC-ROC_Equality}
	  \end{subfigure}
	\hspace{1mm}
	\begin{subfigure}{0.23\textwidth}
		\centering
		\includegraphics[width=\textwidth]{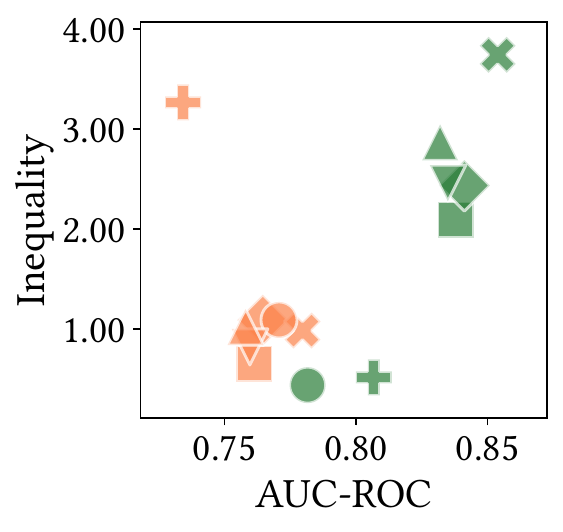}
		\caption{\penn}
		\label{fig:Penn94_AUC-ROC_Equality}
	\end{subfigure}
	\hspace{1mm}
	\begin{subfigure}{0.23\textwidth}
		  \centering
		  \includegraphics[width=\textwidth]{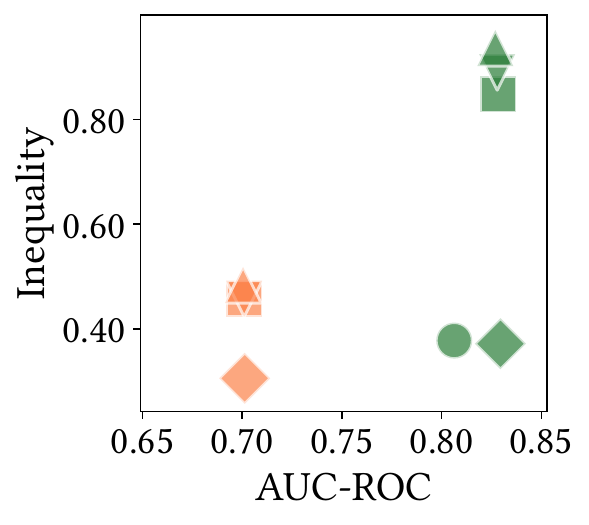}
		  \caption{\pokecz}
		  \label{fig:Pokec-Z_AUC-ROC_Equality}
	  \end{subfigure}
	  \caption{Accuracy (X-axis, larger is better) measured using AUC-ROC versus algorithmic discrimination (Y-axis, smaller is better) measured using Disparity (top row) and Inequality (bottom row) reported as percentages. The optimal is towards the bottom right in all plots which denotes higher AUC-ROC and lower Disparity and Equality.}
	\label{fig:accuracy_fairness_tradeoff}
  \end{figure*}

\subsection{Methodology}\label{subsec:experimental-setup}

\paragraph{Processing Datasets.}
\cikm{\citet{agarwal2021towards} and \citet{dong2022edits} utilize a non-standardized method for creating dataset splits that does not include all nodes. 
Following convention, we create new stratified random splits such that the label imbalance in the original data is reflected in each of the training, validation, and test sets.
For \german, \credit, and \pokecz, we use 60\% of the dataset for training, 20\% for validation, and the remaining 20\% for testing. 
For \penn, we use only 20\% for training and validation (each) because we find that is sufficient for GNNs, with the remaining 60\% used for testing.
Additionally, we adapt the datasets for use by \pfr as described previously (cf. Section~\ref{subsec:pfras}).
\footnote{Unlike ours, \citet{song2022guide} compare with \pfr without employing ranking variables. 
Further, they set both \WF{} and \WX{} to the (normalized) adjacency matrix to fit-transform node attributes which is different from the prescribed specifications by \citet{lahoti2019pfr}.
URL: \url{https://github.com/weihaosong/GUIDE} (retrieved April 2023).}
For computing between-group quantile graphs, we choose Loan Amount, Maximum Bill Amount Over Last 6 Months, Spoken Language, and F6 as ranking variables for \german, \credit, \pokecz, and \penn respectively.
}

\paragraph{Interventions.}
\cikm{Each intervention in our study is benchmarked against the performance of three vanilla GNNs namely, \gcn, \sage, and \gin, referred to as \original.
We construct \pfras to debias \X and \A{} as per Section~\ref{subsec:pfras}.
For ablation, we consider two variants: (i) \pfr-X that only applies \pfr on \X, (ii) \pfr-A that applies only \pfr on a DeepWalk embedding and reconstructs a graph using \asim.}

\cikm{We vary \flipfrac from 0.1 (1\%) to 0.4 (40\%) in increments of 0.1.
For each \flipfrac, we use the same hyperparameters that returned the maximum accuracy for vanilla GNNs and post-process their predictions as per Algorithm~\ref{algo:postprocess}. 
For each seed and \flipfrac, we randomly select \flipfrac fraction of nodes from the protected class with a predicted negative outcome and smoothen over 20 trials.
We define heavy and light versions of \postprocess namely, (i) \textsc{PostProcess}+ and (ii) \textsc{PostProcess}-, in terms of \flipfrac.
\textsc{PostProcess}+ is defined at that value of \flipfrac where disparity is lowest compared to \original and \textsc{PostProcess}- is set halfway between the disparity of \original and \textsc{PostProcess}+.}

\cikm{We compare these with three baselines: 
(i) \unaware (that naively deletes the sensitive attribute column from \X),
(ii) \edits~\cite{dong2022edits}, and
(iii) \nifty~\cite{agarwal2021towards}. %, and
% (iv) \guide~\cite{song2022guide}.
Previous studies do not consider \unaware which is a competitive baseline according to our results (see below).}

\begin{figure*}[ht]
	\centering
	\includegraphics[width=0.65\textwidth]{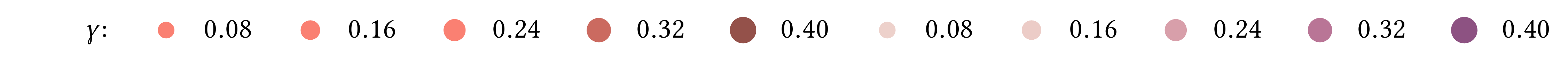}
	\hspace{2mm}
	\\[1pt]

	\begin{subfigure}{0.225\textwidth}
	  \centering
	  \includegraphics[width=\textwidth,clip,trim={0 4mm 0 0}]{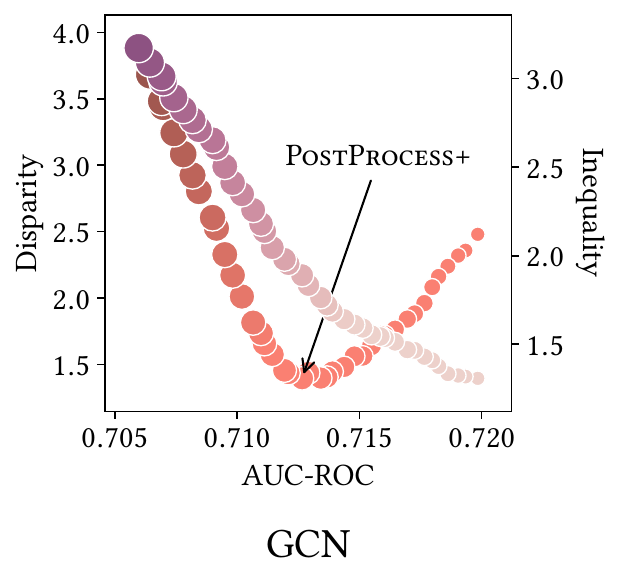}
	  \label{fig:Credit_GCN_postprocess_AUC-ROC}
	\end{subfigure}
	\hspace{5mm}
	\begin{subfigure}{0.225\textwidth}
		\centering
		\includegraphics[width=\textwidth,clip,trim={0 4mm 0 0}]{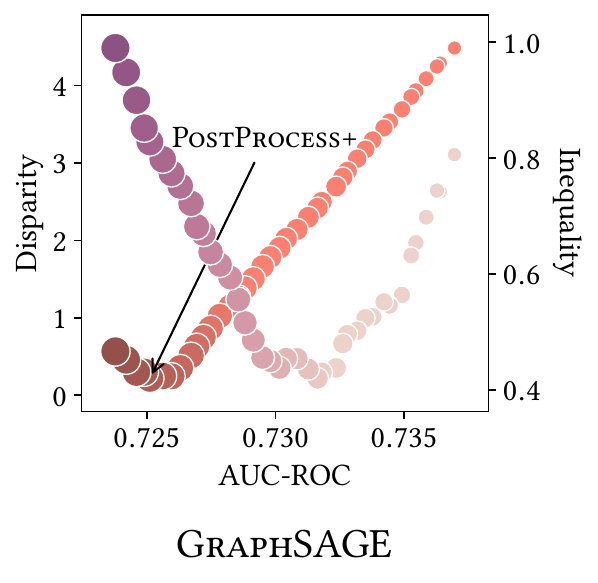}
		\label{fig:Credit_GraphSAGE_postprocess_AUC-ROC}
	\end{subfigure}
	\hspace{5mm}
	\begin{subfigure}{0.225\textwidth}
		\centering
		\includegraphics[width=\textwidth,clip,trim={0 4mm 0 0}]{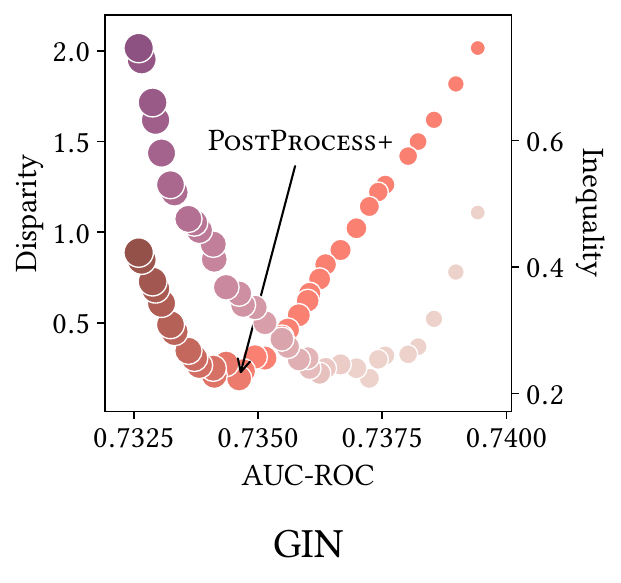}
		\label{fig:Credit_GIN_postprocess_AUC-ROC}
	\end{subfigure}
	
	\caption{AUC-ROC as a function of Disparity (red) and Inequality (purple) for varying levels of the \flipfrac parameter of \postprocess on the \credit dataset. Higher values of \flipfrac are depicted by larger marker shapes and darker colors and indicate heavier interventions. \cikm{As \flipfrac increases, AUC-ROC always decreases and Equality increases. Disparity first decreases upto an inflection point and then increases indicating an over-correction towards the protected class.}}
	\label{fig:postprocess_tradeoff}
  \end{figure*}

\begin{figure*}[ht]
	\centering
	\includegraphics[width=0.65\textwidth]{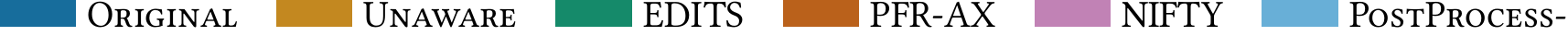}
	\\[2pt]
	\begin{subfigure}{0.315\textwidth}
	  \centering
	  \includegraphics[width=\textwidth]{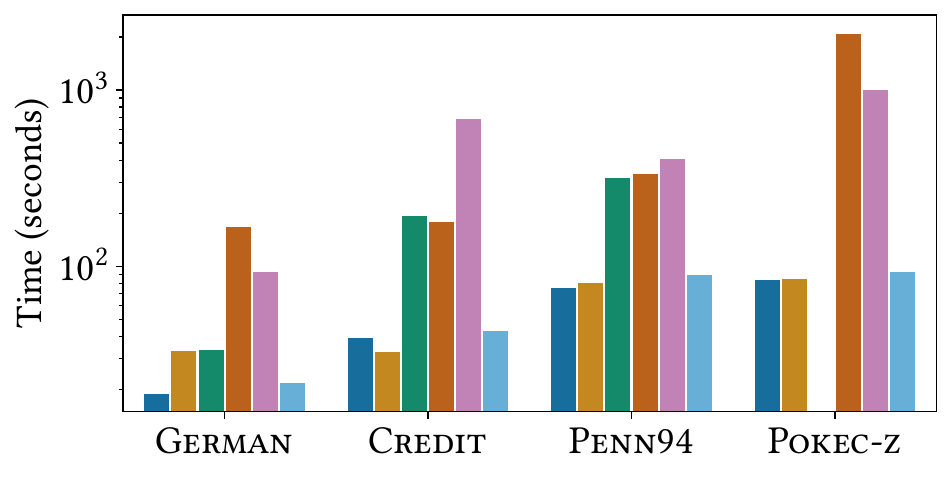}
	  \caption{\gcn}
	  \label{fig:intervention_runtime_GCN}
	\end{subfigure}
	\begin{subfigure}{0.315\textwidth}
		\centering
		\includegraphics[width=\textwidth]{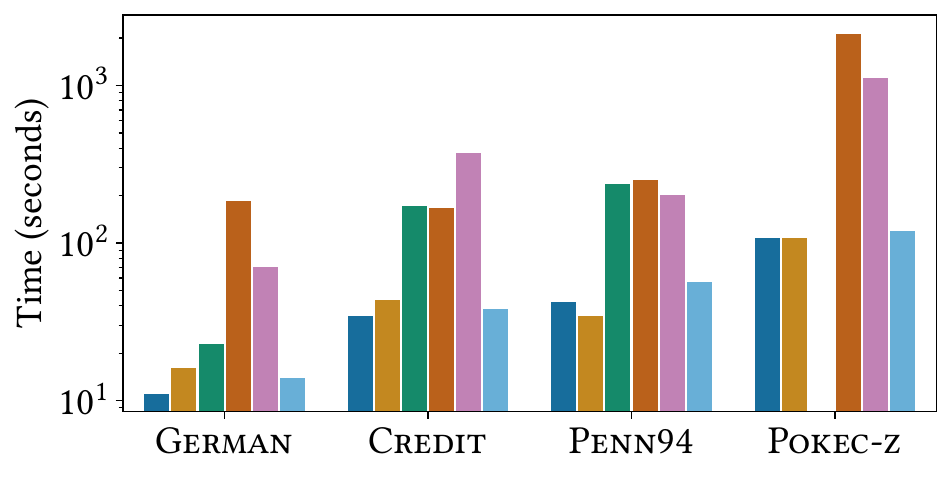}
		\caption{\sage}
		\label{fig:intervention_runtime_GraphSAGE}
	\end{subfigure}  
	\begin{subfigure}{0.315\textwidth}
		\centering
		\includegraphics[width=\textwidth]{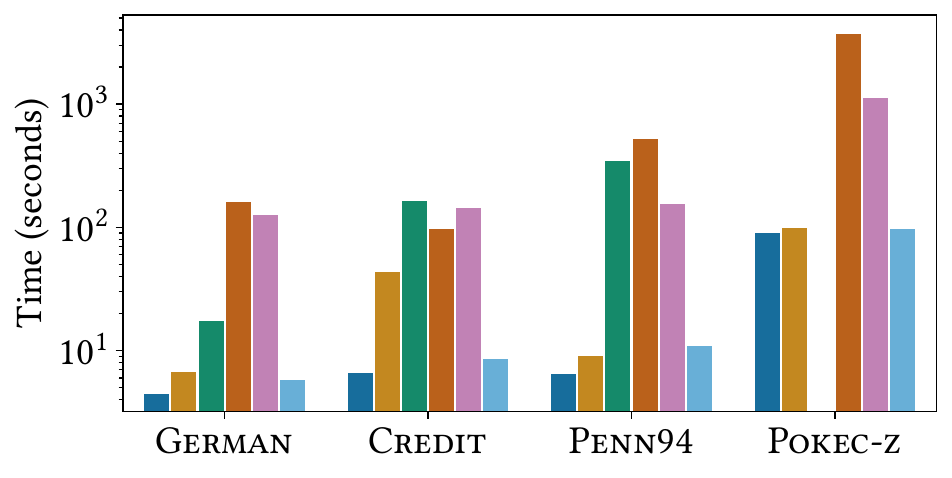}
		\caption{\gin}
		\label{fig:intervention_runtime_GIN}
	\end{subfigure}  
	\caption{Runtime in seconds (log-scale) of various interventions on \gcn, \sage, and \gin for \german, \credit, \penn, and \pokecz increasing with dataset size. \cikm{\postprocess is fastest because updating model inference is inexpensive.}}
	\label{fig:runtime_gnn}
\end{figure*}

\paragraph{Training.}
\cikm{We set $\embdim = 128$ dimensions for \deepwalk.
Depending on the dataset and interventions, we allow models to train for 500, 1000, 1500, or 2000 epochs.
As per convention, we report results for each model/intervention obtained after $T$ epochs and averaged over 5 runs.
This is different from previous studies such as \nifty that train for (say) $T$ epochs and report results for that model instance that has the best validation score from upto $T$ epochs.
This, combined with our stratified splits, is a key factor for observing materially different scores from those reported by the original authors.
To ensure fair comparison, we tune hyperparameters for each intervention and model via a combination of manual grid search and Bayesian optimization using WandB~\cite{wandb}.
The goal of this hyperparameter optimization is to find that setting of hyperparameters that results in a model with a maximal AUC-ROC score while aiming to have lower disparity and equality scores than \original.}

\paragraph{Implementation.}
We implement our models and interventions in Python 3.7.
We use SNAP's C++ implementation for \deepwalk.
\cikm{\edits\footnote{\cikm{\url{https://github.com/yushundong/EDITS} (retrieved April 2022)}} and \nifty\footnote{\cikm{\url{https://github.com/chirag126/nifty} (retrieved April 2022)}} are adapted from their original implementations.}
Our experiments were conducted on a Linux machine with 32 cores, 100 GB RAM, and an V100 GPU.
Our code is available at \url{https://github.com/arpitdm/gnn\_accuracy\_fairness_tradeoff}.

\subsection{Results} \label{subsec:experiment_results}

\paragraph{Algorithmic Fairness-Accuracy Tradeoff.}
Figure~\ref{fig:accuracy_fairness_tradeoff} presents AUC-ROC (X-axis) against disparity (Y-axis) in the first row and inequality (Y-axis) in the second row achieved by various interventions for the four datasets (cf. RQ1). 
We represent the vanilla \gcn model as an orange square and use different orange markers for different interventions on \gcn.
\sage is similarly illustrated in green.
Interventions that cause a $>5\%$ (multiplicative) decrease in AUC-ROC compared to the vanilla model are omitted from the plot.
Since higher values of AUC-ROC and lower values of \parity and \equality are better, the optimal position is towards the bottom right in each plot (cf. RQ2). 
\cikm{For ease of presentation, we defer full results for \gin and all interventions to Table~\ref{tab:full_accuracy_fairness} in Appendix~\ref{app:experiment-results}.}

\cikm{Across datasets, \sage and \gin are more accurate than \gcn but \sage displays higher disparity and inequality while \gin displays lower.
\pfras and \postprocess- offer better tradeoffs than other baselines for \german and \credit across models. 
This translates to upto 70\% and 80\% lower disparity than \original at less than 5\% and 1\% decrease in accuracy on \german, respectively.
In comparison, \nifty offers 60\% lower disparity (2.18\% vs. 5.16\% on \german) at a 4.22\% reduction in AUC-ROC.
The lack of correlation between decreases in disparity and inequality may be explained in part by the impossibility theorem showing that these two criteria cannot be optimized simultaneously~\citet{chouldechova2017fair}.
In \penn and \pokecz, \pfr-A and \pfr-X are more effective than \pfras (cf. Table~\ref{tab:full_accuracy_fairness}). 
We caveat the use of \postprocess in these datasets because choosing nodes randomly displays unintended consequences in maintaining accuracy without promoting fairness.
\unaware proves effective across models and is especially optimal for \pokecz.
\edits proves a heavy intervention causing large reductions in accuracy for relatively small gains in disparity.
}
% In \pokecz, \pfras offers large reductions in disparity (but at larger cost to AUC compared with \nifty while \unaware outperforms other baselines.
% However, in this case, both \pfras and \postprocess offer better tradeoffs--upto 70\% and 80\% lower disparity at less than 5\% and 1\% decrease in accuracy, respectively.
% We find similar results in \equality for \german for nearly all interventions.
% However, this is not the case in \credit where \postprocess and \nifty display reduced disparity compared to \original while their inequality scores are higher. 
% %
% This finding can be explained in part by the impossibility theorem first proven in \citet{chouldechova2017fair} that states that these two criteria cannot be optimized simultaneously.

\begin{figure*}[ht]
	\captionsetup[sub]{skip=3pt}
	\centering
	\includegraphics[width=0.6\textwidth]{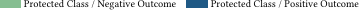}
	\\

	\begin{subfigure}{0.19\textwidth}
		\centering
		\includegraphics[width=\textwidth]{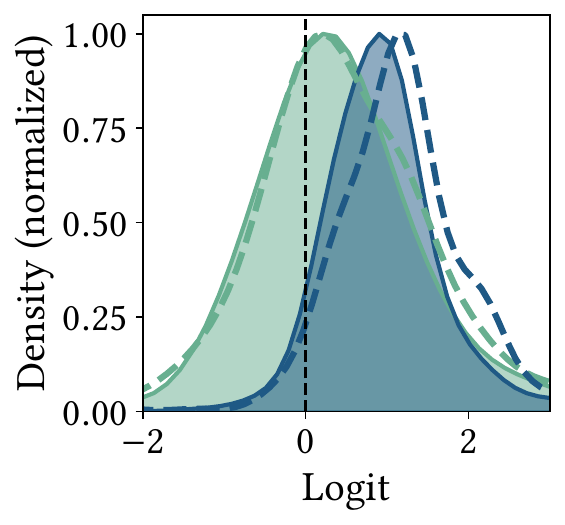}
		\label{fig:Credit_GCN_Unaware_logit_density}
	\end{subfigure}
	\begin{subfigure}{0.19\textwidth}
		\centering
		\includegraphics[width=\textwidth]{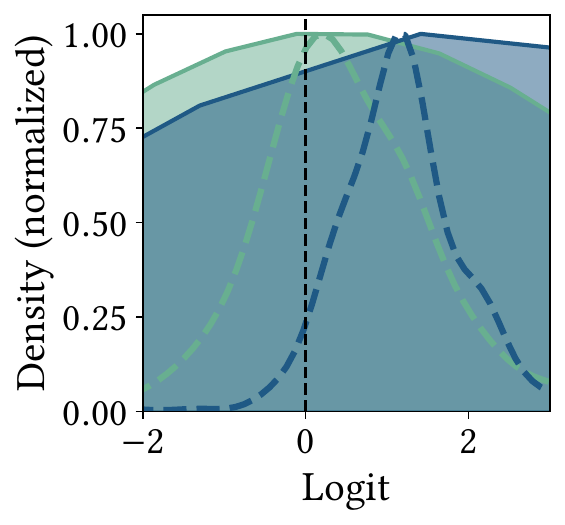}
		\label{fig:Credit_GCN_EDITS_logit_density}
	\end{subfigure}
	\begin{subfigure}{0.19\textwidth}
		\centering
		\includegraphics[width=\textwidth]{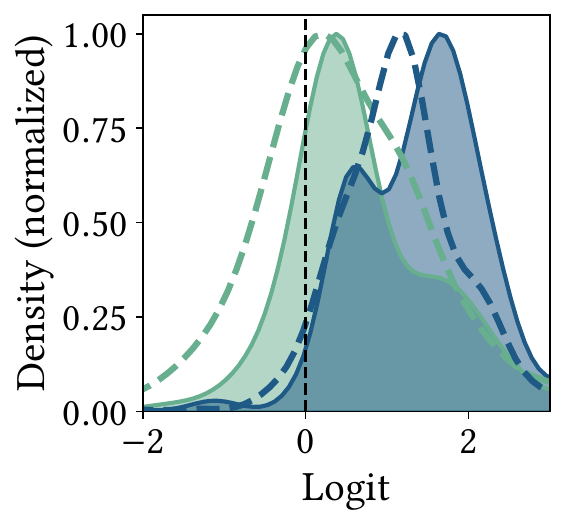}
		\label{fig:Credit_GCN_PFR_logit_density}
	\end{subfigure}
	\begin{subfigure}{0.19\textwidth}
		\centering
		\includegraphics[width=\textwidth]{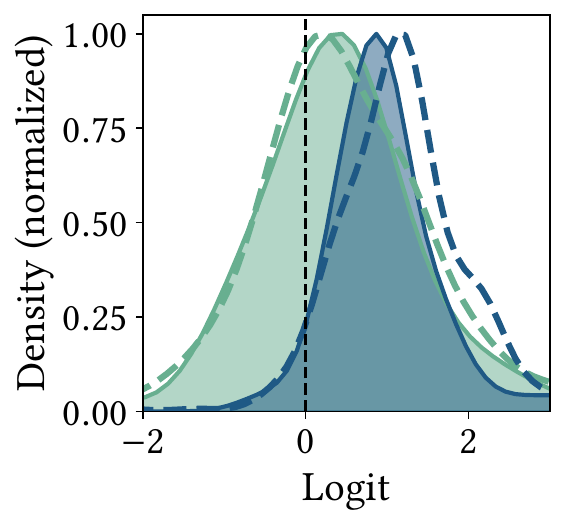}
		\label{fig:Credit_GCN_NIFTY_logit_density}
	\end{subfigure}
	\begin{subfigure}{0.19\textwidth}
		\centering
		\includegraphics[width=\textwidth]{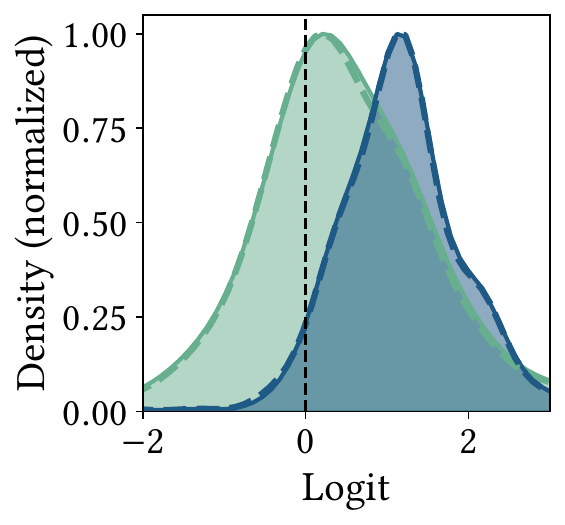}
		\label{fig:Credit_GCN_PostProcess-_logit_density}
	\end{subfigure}

	\begin{subfigure}{0.19\textwidth}
		\centering
		\includegraphics[width=\textwidth]{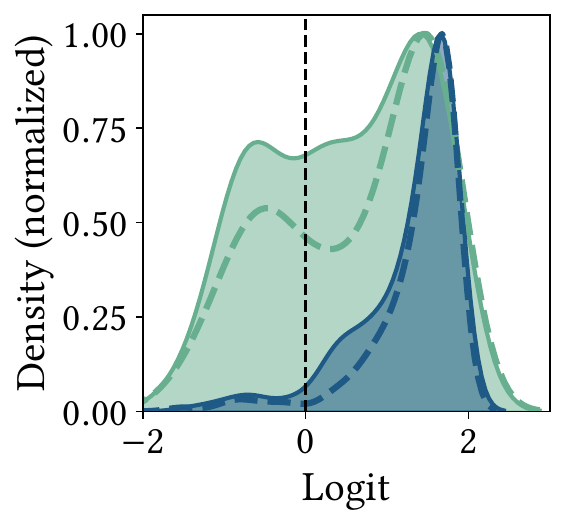}
		\label{fig:Credit_GraphSAGE_Unaware_logit_density}
	  \end{subfigure}
	  \begin{subfigure}{0.19\textwidth}
		\centering
		\includegraphics[width=\textwidth]{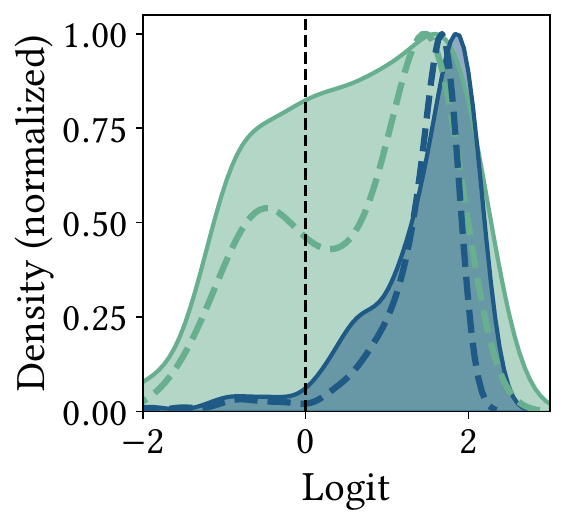}
		\label{fig:Credit_GraphSAGE_EDITS_logit_density}
	  \end{subfigure}
	  \begin{subfigure}{0.19\textwidth}
		\centering
		\includegraphics[width=\textwidth]{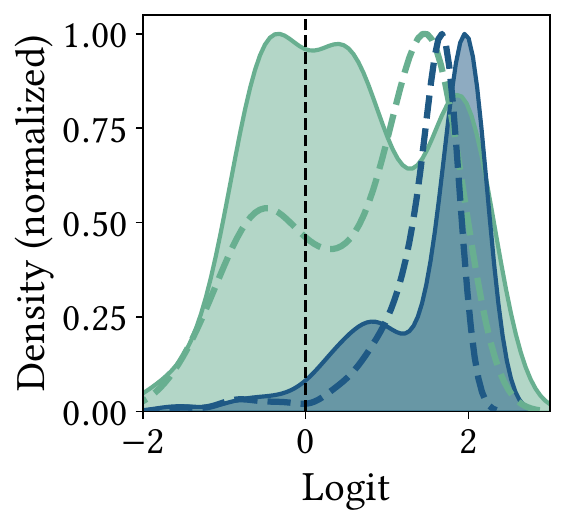}
		\label{fig:Credit_GraphSAGE_PFR_logit_density}
	  \end{subfigure}
	  \begin{subfigure}{0.19\textwidth}
		\centering
		\includegraphics[width=\textwidth]{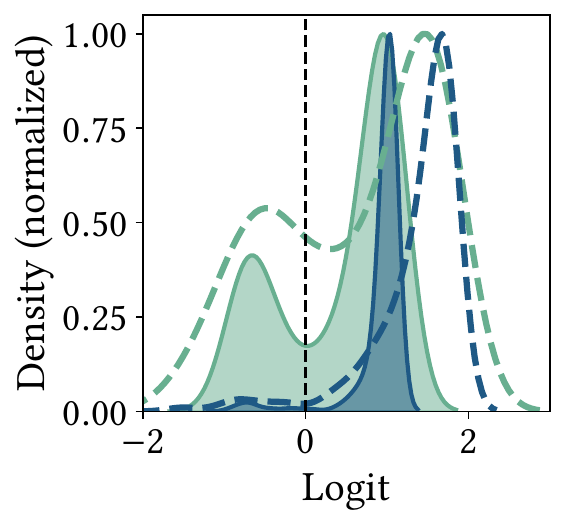}
		\label{fig:Credit_GraphSAGE_NIFTY_logit_density}
	  \end{subfigure}
	  \begin{subfigure}{0.19\textwidth}
		\centering
		\includegraphics[width=\textwidth]{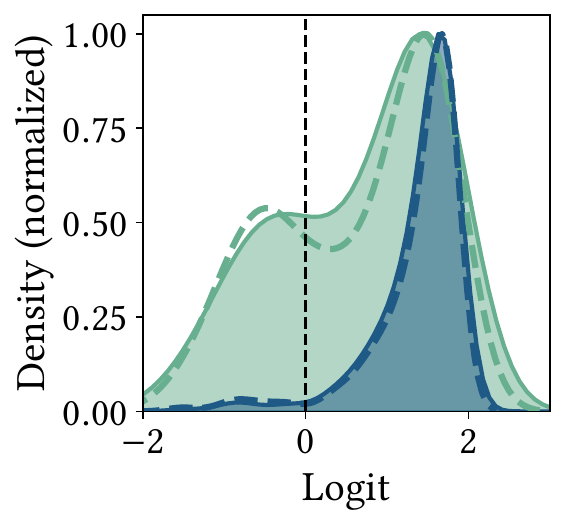}
		\label{fig:Credit_GraphSAGE_PostProcess-_logit_density}
	  \end{subfigure}

	  \begin{subfigure}{0.19\textwidth}
		\centering
		\includegraphics[width=\textwidth]{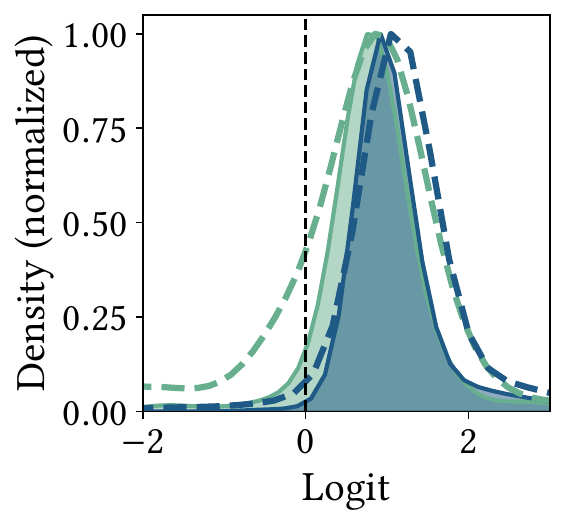}
		\caption{\unaware}
		\label{fig:Credit_GIN_Unaware_logit_density}
	  \end{subfigure}
	  \begin{subfigure}{0.19\textwidth}
		\centering
		\includegraphics[width=\textwidth]{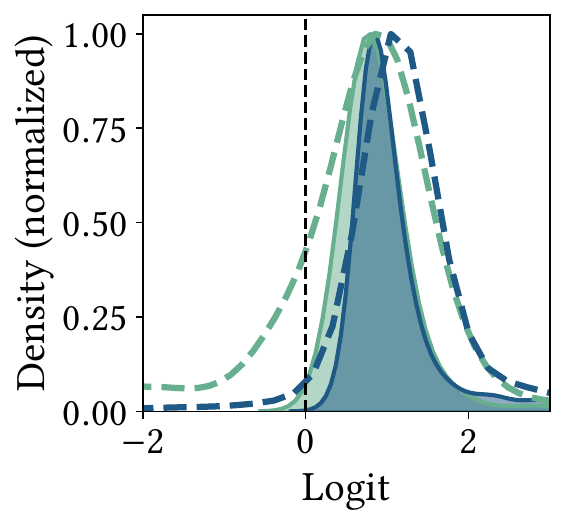}
		\caption{\edits}
		\label{fig:Credit_GIN_EDITS_logit_density}
	  \end{subfigure}
	  \begin{subfigure}{0.19\textwidth}
		\centering
		\includegraphics[width=\textwidth]{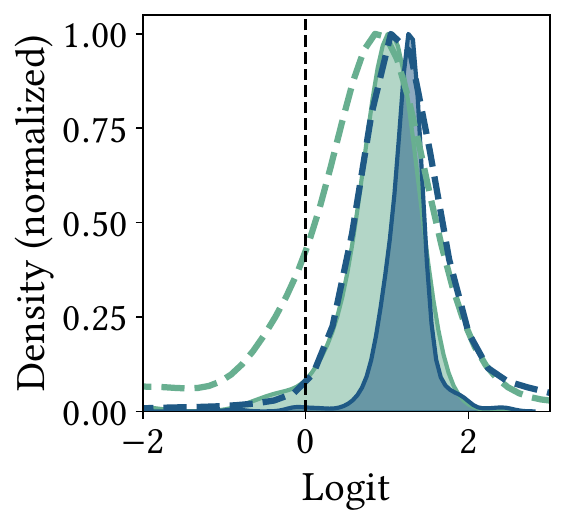}
		\caption{\pfras}
		\label{fig:Credit_GIN_PFR_logit_density}
	  \end{subfigure}
	  \begin{subfigure}{0.19\textwidth}
		\centering
		\includegraphics[width=\textwidth]{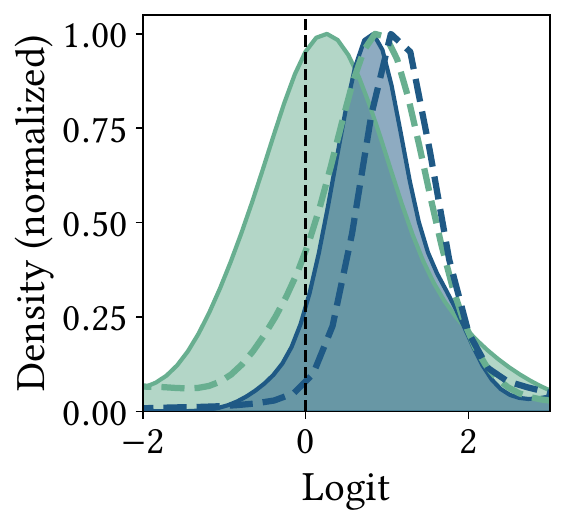}
		\caption{\nifty}
		\label{fig:Credit_GIN_NIFTY_logit_density}
	  \end{subfigure}
	  \begin{subfigure}{0.19\textwidth}
		\centering
		\includegraphics[width=\textwidth]{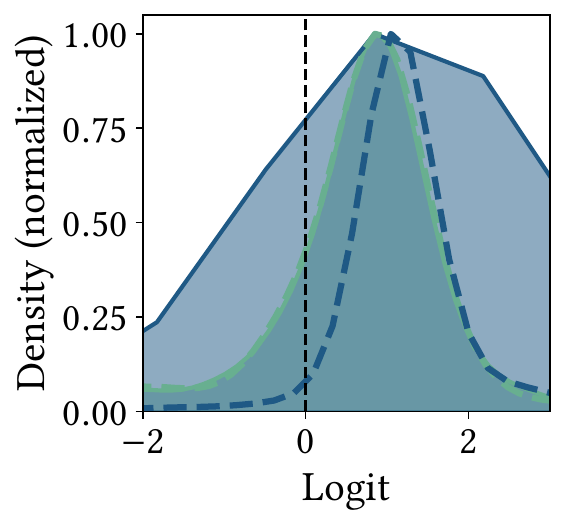}
		\caption{\postprocess-}
		\label{fig:Credit_GIN_PostProcess-_logit_density}
	  \end{subfigure}

		\caption{Density of logit scores of \gcn (first row), \sage (second row), and \gin (third row) after applying different algorithmic fairness interventions for users in the protected class in the \credit dataset. The vertical dashed (black) line depicts the threshold used for label prediction (positive scores indicate positive outcomes). The colored dashed curves indicate the density of output scores of \original. 
		\cikm{\pfras and \postprocess- improve model confidence (density) for correctly predicting a positive label for users in the protected class.}
		}
	\label{fig:Credit_model_logit_kernel_density}
  \end{figure*}

\paragraph{Sensitivity to \flipfrac.}
Figure~\ref{fig:postprocess_tradeoff} trades off AUC (X-axis), disparity (left Y-axis, red points), and inequality (right Y-axis, purple points) for \gcn, \sage, and \gin on \credit as a function of \flipfrac. 
Due to large label imbalance in \credit and small number of nodes with negative predicted outcomes from the protected class, varying \flipfrac by 1\% translates to changing predictions for 7 nodes.
\postprocess thus offers granular control.
As \flipfrac increases, AUC-ROC decreases while \parity first reduces and then increases again.
This inflection point indicates that the post-processing policy is overcorrecting in favour of the protected class resulting in disparity towards the non-protected class.
Conversely, such improvements are absent in \pokecz since vanilla GNNs themselves are inherently less biased.

\paragraph{Runtime.}
\cikm{Figure~\ref{fig:runtime_gnn} depicts the total computation time in seconds (on log-scale) for each intervention on the four datasets for \gcn, \sage, and \gin.
We observe similar trends for all three GNN models.
As expected, the larger the dataset, the higher the runtime.}
Updating a model's predictions at inference time is inexpensive and the resulting overhead for \postprocess is thus negligible.
The running time for \pfras increases significantly with increasing dataset size.
The key bottlenecks are very eigenvalue decompositions for sparse, symmetric matrices in \pfr requiring $\bigObound\round{\numnodes^3}$ time and constructing DeepWalk embeddings. 
For instance, in the case of \pokecz, \pfr required (on average) 47 minutes in our tests while \asim and GNN training required less than 5 minutes each.
For comparison, \nifty required approximately 22 minutes while \edits did not complete due to memory overflow.

\paragraph{Model Confidence.}
In Figure~\ref{fig:Credit_model_logit_kernel_density}, we display the impact of fairness interventions on a model's confidence about its predictions, i.e., uncalibrated density (Y-axis), compared to its logit scores (X-axis) on the \credit dataset.
The plots in the top, middle, and bottom rows corresponds to \gcn, \sage, and \gin, respectively.
Larger positive values imply higher confidence about predicting a positive outcome and larger negative values imply higher confidence for a negative outcome prediction.
While there isn't a universal desired outcome, an intermediate goal for an intervention may be to ensure that a model is equally confident about \emph{correctly} predicting both positive and negative labels.
Blue regions show normalized density of logit values for nodes in the protected class with a positive ground-truth label (\texttt{S1-Y1}) and green regions show the same for nodes in the protected class with a negative outcome as ground-truth.
The dashed colored lines indicate density values for these groups of nodes for the \original model.
\postprocess and \unaware induce small changes to GNN's outputs while \edits is significantly disruptive. 
\pfras nudges the original model's output for nodes in \texttt{S1-Y1} away from 0 making it more confident about its positive (correct) predictions while \nifty achieves the reverse.

\section{Conclusion}
\label{sec:conclusion}

We presented two interventions that intrinsically differ from existing methods:
\pfras debiases data prior to training to connect similar nodes across protected and non-protected groups while seeking to preserve existing degree distributions;
\postprocess updates model predictions to reduce error rates across protected user groups. 
We frame our study in the context of the tension between disparity, inequality, and accuracy and quantify the scope for improvements and show that our approaches offer intuitive control over this tradeoff.
Given their model-agnostic nature, we motivate future analysis by combining multiple interventions at different loci in the learning pipeline.

\begin{acks}
	This work has been partially supported by:
	Department of Research and Universities of the Government of Catalonia (SGR 00930),
	EU-funded projects ``SoBigData++'' (grant agreement 871042), ``FINDHR'' (grant agreement 101070212)
	and MCIN/AEI /10.13039/501100011033 under the Maria de Maeztu Units of Excellence Programme (CEX2021-001195-M).
	We also thank the reviewers for their useful comments.
\end{acks}  

%%%%%%%%%%%%%%%%%%%%%%%%%%%%%%%%%%%%%%%%%%%%%%%%%%%%%%%%%%%%%%%%%%%%%%%%%%%%%%%
% APPENDIX
%%%%%%%%%%%%%%%%%%%%%%%%%%%%%%%%%%%%%%%%%%%%%%%%%%%%%%%%%%%%%%%%%%%%%%%%%%%%%%%
% \newpage
\clearpage

\appendix

\twocolumn[\section{Additional Experimental Results}\label{app:experiment-results}

	\vskip\intextsep

	\noindent\begin{minipage}{\textwidth}

	\centering  

	\fontsize{8.2}{14}\selectfont
	\captionof{table}{Accuracy (AUC-ROC) and algorithmic fairness (Disparity and Inequality) scores for 8 interventions for \textsc{GCN}, \textsc{GraphSAGE}, and \textsc{GIN} models on four datasets. Results are averaged across five runs. Higher values of AUC (fraction between 0 and 1) indicate higher performance. Lower values of disparity ($\Delta_{\text{SP}}$) and inequality ($\Delta_{\text{EO}}$) in percentage indicate higher algorithmic fairness. No single intervention dominates all others across datasets and models. However, \textsc{PostProcess-} generally offers a gentle accuracy-fairness tradeoff. A dashed line denotes out-of-memory.}
	\label{tab:full_accuracy_fairness} 

	\vspace{-1em}

	\begin{tabular}{cccccccccccc} 

	\toprule 

		\textbf{Dataset} & \textbf{Model}     &  \textbf{Metric}         &  \textsc{Original} &  \textsc{Unaware} &  \textsc{EDITS} &  \textsc{PFR}-\textsc{A} &  \textsc{PFR}-\textsc{X} &  \textsc{PFR}-\textsc{AX} &  \textsc{NIFTY} &  \textsc{PostProcess+} &  \textsc{PostProcess-} \\
	\midrule 

	\multirow{9}{*}{German} & \multirow{3}{*}{GCN} & AUC-ROC &              0.687 &             0.688 &                  0.695 &                   0.643 &                   0.698 &                  0.638 &           0.658 &                  0.670 &                  0.682 \\
         &     & Parity &              5.156 &             2.878 &                  3.625 &                   2.068 &                   4.204 &                  3.878 &           2.182 &                  1.082 &                  3.250 \\
         &     & Equality &              1.260 &             1.690 &                  2.215 &                    1.54 &                   2.612 &                  2.112 &           3.094 &                  3.668 &                  2.242 \\
\cline{2-12}
         & \multirow{3}{*}{GraphSAGE} & AUC-ROC &              0.688 &             0.685 &                  0.691 &                   0.665 &                   0.708 &                  0.708 &           0.653 &                  0.680 &                  0.685 \\
         &     & Parity &              4.450 &             5.034 &                  4.582 &                   2.856 &                   3.072 &                  2.758 &           3.976 &                  0.970 &                  2.864 \\
         &     & Equality &              3.974 &             3.748 &                  3.566 &                   2.804 &                   5.174 &                  4.348 &           3.036 &                  1.482 &                  2.576 \\
\cline{2-12}
         & \multirow{3}{*}{GIN} & AUC-ROC &              0.709 &             0.707 &                  0.675 &                   0.664 &                   0.619 &                   0.59 &           0.654 &                  0.680 &                  0.696 \\
         &     & Parity &              8.600 &             1.496 &                   5.61 &                   2.714 &                   1.218 &                  2.602 &           2.118 &                  2.882 &                  6.058 \\
         &     & Equality &              2.168 &             4.260 &                  2.824 &                     1.6 &                   1.362 &                  4.476 &           3.278 &                  2.216 &                  1.624 \\
\cline{1-12}
\cline{2-12}
\multirow{9}{*}{Credit} & \multirow{3}{*}{GCN} & AUC-ROC &              0.720 &             0.681 &                  0.704 &                   0.721 &                   0.735 &                  0.727 &           0.715 &                  0.713 &                  0.718 \\
         &     & Parity &              2.518 &             6.194 &                   2.12 &                   3.094 &                   0.316 &                  1.344 &           3.614 &                  1.396 &                  1.962 \\
         &     & Equality &              1.332 &             4.550 &                  0.944 &                   1.274 &                   0.444 &                  0.762 &           0.610 &                  1.890 &                  1.430 \\
\cline{2-12}
         & \multirow{3}{*}{GraphSAGE} & AUC-ROC &              0.737 &             0.739 &                  0.744 &                    0.72 &                   0.751 &                  0.747 &           0.726 &                  0.725 &                  0.731 \\
         &     & Parity &              4.484 &             3.876 &                    3.9 &                   3.866 &                   2.746 &                   2.67 &           4.112 &                  0.218 &                  2.298 \\
         &     & Equality &              0.806 &             0.302 &                  0.276 &                   1.184 &                   0.366 &                   0.98 &           1.042 &                  0.828 &                  0.436 \\
\cline{2-12}
         & \multirow{3}{*}{GIN} & AUC-ROC &              0.739 &             0.713 &                  0.707 &                   0.724 &                   0.742 &                  0.716 &           0.716 &                  0.735 &                  0.737 \\
         &     & Parity &              2.016 &             0.686 &                   0.86 &                    1.32 &                   0.612 &                  1.616 &           3.268 &                  0.194 &                  1.142 \\
         &     & Equality &              0.486 &             0.296 &                  0.494 &                   1.204 &                   0.776 &                  0.416 &           0.440 &                  0.358 &                  0.224 \\
\cline{1-12}
\cline{2-12}
\multirow{9}{*}{Penn94} & \multirow{3}{*}{GCN} & AUC-ROC &              0.761 &             0.765 &                   0.78 &                    0.69 &                   0.796 &                  0.734 &           0.771 &                  0.758 &                  0.760 \\
         &     & Parity &              1.858 &             1.806 &                  2.208 &                   1.856 &                    1.21 &                   1.44 &           1.014 &                  2.820 &                  2.340 \\
         &     & Equality &              0.650 &             1.086 &                  0.982 &                   3.046 &                   2.254 &                  3.264 &           1.088 &                  1.016 &                  0.818 \\
\cline{2-12}
         & \multirow{3}{*}{GraphSAGE} & AUC-ROC &              0.838 &             0.841 &                  0.854 &                   0.732 &                    0.83 &                  0.807 &           0.782 &                  0.832 &                  0.835 \\
         &     & Parity &              3.762 &             3.908 &                   5.09 &                    2.54 &                   4.326 &                  4.734 &           1.998 &                  5.154 &                  4.456 \\
         &     & Equality &              2.090 &             2.432 &                   3.74 &                   0.922 &                   1.692 &                  0.514 &           0.436 &                  2.864 &                  2.456 \\
\cline{2-12}
         & \multirow{3}{*}{GIN} & AUC-ROC &              0.789 &             0.778 &                  0.776 &   0.717 &   0.694 &                  0.731 &           0.769 &                  0.784 &                  0.786 \\
         &     & Parity &              1.546 &             1.058 &                  1.474 &   1.081 &   1.670 &                   1.182 &           0.862 &                  3.116 &                  2.320 \\
         &     & Equality &              2.358 &             1.732 &                  3.306 &   4.584 &   2.672 &                   2.368 &           1.966 &                  1.130 &                  1.732 \\
\cline{1-12}
\cline{2-12}
\multirow{9}{*}{Pokec-Z} & \multirow{3}{*}{GCN} & AUC-ROC &              0.701 &             0.701 &  -- &                    0.66 &                   0.616 &                  0.615 &           0.627 &                  0.700 &                  0.701 \\
         &     & Parity &              0.530 &             0.378 &  -- &                   0.235 &                   0.216 &                  0.502 &           0.700 &                  0.582 &                  0.530 \\
         &     & Equality &              0.458 &             0.306 &  -- &                   0.412 &                   0.078 &                  0.486 &           0.582 &                  0.484 &                  0.458 \\
\cline{2-12}
         & \multirow{3}{*}{GraphSAGE} & AUC-ROC &              0.828 &             0.830 &  -- &                   0.827 &                   0.775 &  0.778 &           0.806 &                  0.827 &                  0.828 \\
         &     & Parity &              0.860 &             0.384 &  -- &                   0.954 &                    0.33 &  0.327 &           0.374 &                  1.014 &                  0.928 \\
         &     & Equality &              0.848 &             0.372 &  -- &                   0.634 &                   0.464 &  0.458 &           0.378 &                  0.936 &                  0.890 \\
\cline{2-12}
         & \multirow{3}{*}{GIN} & AUC-ROC &              0.712 &             0.710 &  -- &   0.651 &   0.623 &                  0.641 &           0.669 &                  0.712 &                  0.712 \\
         &     & Parity &              0.406 &             0.136 &  -- &   0.721 &   0.181 &                   0.673 &           0.156 &                  0.456 &                  0.406 \\
         &     & Equality &              0.398 &             0.076 &  -- &   1.898 &   1.439 &                    1.887 &           0.114 &                  0.438 &                  0.398 \\
 \bottomrule 
 \end{tabular} 
 \end{minipage} 
 \par\vskip\intextsep]

\clearpage

%%%%%%%%%%%%%%%%%%%%%%%%%%%%%%%%%%%%%%%%%%%%%%%%%%%%%%%%%%%%%%%%%%%%%%%%%%%%%%%%
% References
%%%%%%%%%%%%%%%%%%%%%%%%%%%%%%%%%%%%%%%%%%%%%%%%%%%%%%%%%%%%%%%%%%%%%%%%%%%%%%%%

\bibliographystyle{ACM-Reference-Format}
\balance
\bibliography{references}

\end{document}